\documentclass[cameraready]{Interspeech}

\title{Unlocking In-Context Learning in Audio-Language Models from Decentralized Medical Audio}
\author[affiliation=1]{Ran}{Piao}
\author[affiliation=1]{Tsai-Ning}{Wang}
\author[affiliation=2]{Martijn}{den Dekker}
\author[affiliation=3]{Linda}{Moonen}
\author[affiliation=4]{Hareld}{Kemps}
\author[affiliation=1]{Yuan}{Lu}
\author[affiliation=1]{Aaqib}{Saeed}

\address{
  $^1$Eindhoven University of Technology, The Netherlands \\
  $^2$Erasmus MC, The Netherlands \\
  $^3$Rijnstate Hospital, The Netherlands \\
  $^4$M\'{a}xima MC Hospital, The Netherlands
}
\email{r.piao@tue.nl}

\usepackage{amsmath}
\usepackage{comment}
\usepackage{multirow}
\usepackage{subcaption}
\usepackage{adjustbox}
\usepackage{float}
\usepackage{dblfloatfix}
\usepackage{mdframed}
\usepackage{xcolor}
\usepackage{colortbl}
\usepackage{placeins}

\begin{document}

\maketitle

\begin{abstract}
Clinical audio diagnosis in low-resource settings requires models that identify conditions from minimal examples without large annotated corpora. We propose Federated Self-Contextualization (FSC), a multimodal language model framework for in-context clinical audio diagnosis across federated hospital clients. FSC constructs pseudo-label episodes via unsupervised clustering of audio representations, bypassing scarce real diagnostic labels, and enables contextual reasoning from support–query pairs. Our progressive three-stage pipeline first aligns audio embeddings with the language model via caption-based pretraining, then adapts it for episodic in-context inference through federated optimization. At test time, given a small labeled support set, the model diagnoses an unseen query through multimodal reasoning. On held-out respiratory and cardiac conditions, FSC achieves 71.6\% accuracy in 2-way 2-shot evaluation, outperforming audio-language baselines by over 9\%.
\index{in-context learning, audio-language models, federated learning,
clinical audio diagnosis, healthcare}
\end{abstract}

\section{Introduction}

Clinical audio diagnosis is fundamentally an act of grounded reasoning: a clinician interprets what they \emph{hear}—a wheeze, a murmur, an abnormal cough—in light of what they \emph{know} about the clinical conditions those sounds signify \cite{troncoso2021non,potes2016ensemble}. This process is open-ended, contextual, and knowledge-intensive. Yet the dominant machine learning paradigm for clinical audio reduces it to closed-set classification over a fixed label taxonomy \cite{imran2020ai4covid,chen2021deep,baur2024hear}, discarding both the open-vocabulary nature of real diagnosis and the role of clinical knowledge in interpreting acoustic evidence. Such learning approaches require large centrally aggregated labeled datasets, cannot accommodate diagnostic concepts absent from training, and must be retrained whenever clinical needs evolve—assumptions that rarely hold in practice, where annotated recordings are scarce, distributed across institutional silos, and governed by strict privacy constraints \cite{xu2021federated,rieke2020future}.

Medical language models offer a fundamentally different foundation. Models such as MedGemma \cite{sellergren2025medgemma} have internalized extensive clinical knowledge during pretraining and exhibit a native capacity for in-context learning (ICL): given a few input--output demonstrations, they can adapt to new tasks without parameter updates \cite{dong2024survey}. If an audio signal can be projected into a language model's representational space, these properties become directly available for diagnosis—the model can interpret acoustic evidence against \emph{any} clinical concept expressible in natural language, conditioned on just a few support examples. This makes language models uniquely suited to open-vocabulary clinical audio diagnosis, a capability that purely acoustic few-shot methods such as prototypical networks \cite{wang2020generalizing} cannot provide, since they operate over embedding distances without access to the semantic structure of clinical descriptions.

Grounding a language model in clinical audio, however, poses a central paradox. Training audio-conditioned ICL requires structured episodes of audio--label demonstrations \cite{wu2023large,elizalde2023clap}, yet expert-annotated audio--label pairs are precisely what is unavailable across privacy-constrained clinical institutions. We resolve this paradox through a key insight: \emph{the reasoning skill of contextual diagnosis and the clinical knowledge needed to apply it can be acquired from entirely separate sources.} We propose \textbf{Federated Self-Contextualization (FSC)}, a framework in which a pretrained medical audio encoder \cite{wang2025careaqa} is coupled with a medical language model and trained on episodes constructed from clustering-derived pseudo-labels bearing intentionally meaningless identifiers (e.g., ``Mountain Breeze,'' ``Sun Ray''). Because these pseudo-labels carry no medical semantics, the model cannot memorize disease--sound associations; the only transferable capability it can acquire is the \emph{abstract skill} of mapping acoustic patterns to contextual label descriptions. At evaluation time, when real clinical descriptions such as ``Wheeze'' or ``Atrial Septal Defect'' appear as support labels, the language model's pretrained medical knowledge provides the semantic grounding that bridges the gap—it already understands what these conditions mean, and the learned ICL skill tells it how to match acoustic evidence to the appropriate description.

All training in FSC is performed under a federated protocol \cite{li2020federated,antunes2022federated}, where only model parameters are shared across institutions and patient audio never leaves its source site. Our progressive three-stage pipeline first aligns audio and text representations, then instills episodic in-context reasoning through the language model's lightweight LoRA adapters \cite{hu2022lora}. We evaluate FSC across seven respiratory and cardiac datasets comprising over 22{,}000 recordings. In the 2-way 2-shot setting, FSC achieves 71.6\% diagnostic accuracy, surpassing the strongest audio--language baseline by over 9 percentage points. Our contributions are:
\begin{figure*}[t]
  \centering
  \includegraphics[width=\textwidth]{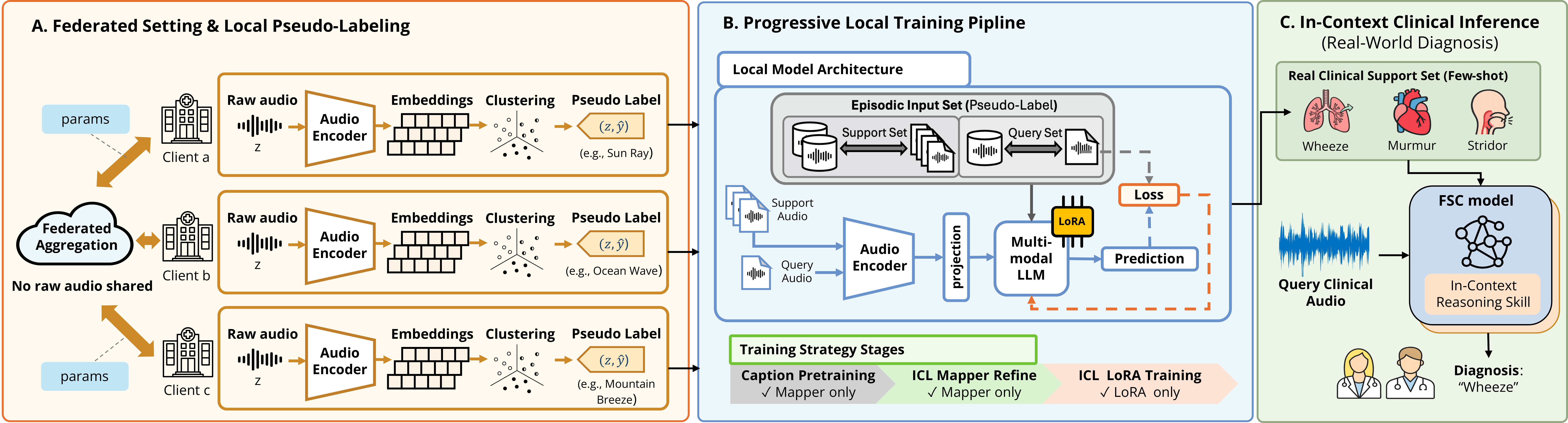}
  \vspace{-0.5cm}
\caption{Overview of Federated Self-Contextualization (FSC). 
\textbf{(A)}~Federated setting with client-side clustering for local pseudo-label generation. 
\textbf{(B)}~Progressive Local ICL Training Pipeline 
\textbf{(C)}~In-context clinical inference using real-world few-shot support examples.}

  \label{fig:framework}
  \vspace{-0.5cm}
\end{figure*}
\begin{itemize}
    \item We formalize \emph{federated self-contextualization}, a paradigm in which semantically void pseudo-labels teach an audio--language model the transferable skill of in-context diagnostic reasoning, while the language model's pretrained clinical knowledge provides the semantic grounding needed to generalize to real diagnostic descriptions at inference; decoupling reasoning skill acquisition from annotation availability.
    \item We instantiate this paradigm under federated constraints through a progressive episodic training pipeline that addresses cross-modal alignment, in-context learning, and data sovereignty jointly, without centralized data access or predefined label spaces.
    \item We demonstrate through extensive evaluation on diverse cardiopulmonary tasks that FSC consistently and substantially outperforms recent audio--language models, with ablations confirming the necessity of each component.
\end{itemize}

\section{Methodology}

\subsection{Problem Formulation}
\vspace{-0.1cm}
We consider clinical audio diagnosis where recordings are distributed across $M$ institutions without centralized access and no expert labels are available during training. Each diagnostic episode $\mathcal{E} = \{\mathcal{S}, x_q\}$ consists of a support set $\mathcal{S} = \{(x_i, y_i)\}_{i=1}^{NK}$ with $K$ audio-label pairs for each of $N$ diagnostic concepts, and a query sample $x_q$. The model predicts:
\[
\hat{y}_q = \arg\max_{y \in \mathcal{Y}_{\mathcal{E}}} \; p_{\text{LM}}(y \mid \mathcal{S}, x_q),
\]
where $\mathcal{Y}_{\mathcal{E}} = \{y_1, \ldots, y_N\}$ are the episode's label descriptions. During training, labels are self-supervised pseudo-labels derived from audio clustering; at evaluation, they are real clinical descriptions of conditions never seen during training.
\vspace{-0.1cm}
\subsection{Model Architecture}
\vspace{-0.1cm}

Each client operates an audio-language model comprising an audio encoder, a projection layer, and a language model backbone (Figure~\ref{fig:framework}B).

The language model backbone is MedGemma-4B-IT~\cite{sellergren2025medgemma}, a 4-billion-parameter instruction-tuned medical language model. MedGemma serves two roles in FSC: its pretraining on biomedical corpora provides the clinical knowledge needed to interpret real diagnostic descriptions at inference, bridging the semantic gap between the void pseudo-labels used during training and medical terminology encountered at evaluation; and its native multimodal input support through visual boundary tokens allows non-textual embeddings to be inserted directly into the token sequence and processed jointly with text via standard self-attention.

The audio encoder is CaReAQA~\cite{wang2025careaqa}, a medical audio encoder pretrained on diverse clinical audio tasks. Given a sample $x$, the encoder produces an embedding $e = E(x) \in \mathbb{R}^{1280}$. A linear projection maps this to a sequence of $L{=}4$ prefix tokens $P(e) \in \mathbb{R}^{L \times d}$, where $d$ is the language model's hidden dimension. These tokens are injected between MedGemma's visual boundary tokens, enabling audio-conditioned generation through the existing self-attention mechanism without any architectural modification to the language model.

\vspace{-0.1cm}
\subsection{Self-Supervised Pseudo-Labeling}
\vspace{-0.1cm}

Within each client $m$, all audio embeddings from $\mathcal{D}_m$ are clustered into $C{=}10$ groups via K-means, and each cluster is assigned a semantically neutral identifier (e.g., ``Mountain Breeze,'' ``Ocean Wave'') deliberately chosen to carry no medical connotation. These assignments serve as pseudo-labels for constructing training episodes. 

The use of semantically void identifiers is a deliberate design choice. Because the pseudo-labels encode no clinical information, the model cannot memorize disease-sound associations; the only transferable capability it can acquire is the abstract reasoning pattern of matching acoustic input to contextual label descriptions. At evaluation, when identifiers are replaced by real descriptions such as ``Wheeze'' or ``Atrial Septal Defect,'' the language model's pretrained medical knowledge grounds these terms semantically, while the learned reasoning skill handles the acoustic comparison.
\vspace{-0.2cm}
\subsection{In-Context Episodic Input Format}
\label{sec:episodic_format}
\vspace{-0.1cm}
Stages~II and~III of training, as well as all inference, operate over episodic inputs. An $N$-way $K$-shot episode is serialized as an interleaved sequence of audio prefix tokens and label text. Each support example appears as audio tokens enclosed in visual boundary tokens\footnote{MedGemma's \texttt{<start\_of\_image>} and \texttt{<end\_of\_image>} tokens.} followed by the label text; the query audio is appended last, and the model is trained to generate the correct label via cross-entropy loss (Figure~\ref{fig:framework}B).

During training, episodes use pseudo-labels from clustering. At inference, real clinical labels are substituted in the same format without any architectural change (Figure~\ref{fig:framework}C). This seamless substitution is precisely where the language model's pretrained clinical knowledge becomes operative: the model recognizes medical terms through its existing representations while applying the in-context reasoning pattern acquired from pseudo-label training.
\vspace{-0.1cm}
\subsection{Progressive Training Pipeline}
\label{sec:training_pipeline}

FSC trains in three stages that progressively build in-context diagnostic capability, each updating a different parameter subset (Table~\ref{tab:training_stages}).

\begin{table}[t]
\centering
\vspace{-0.2cm}
\caption{Training pipeline configuration.}
\label{tab:training_stages}
\footnotesize
\setlength{\tabcolsep}{3pt}
\renewcommand{\arraystretch}{1.1}
\vspace{-0.2cm}
\begin{tabular}{lccc}
\toprule
\textbf{Stage} & \textbf{Format} & \textbf{Trainable} & \textbf{Frozen} \\
\midrule
I: Alignment         & Non-episodic & Proj., Enc. & LM \\
II: Episodic Refine. & Episodic     & Proj., Enc. & LM \\
III: LM Adaptation   & Episodic     & LM (LoRA)   & Proj., Enc. \\
\bottomrule
\end{tabular}
\vspace{-0.55cm}
\end{table}

Stage~I establishes cross-modal alignment by training the encoder and projection on individual audio-label pairs in a non-episodic classification format, with the language model frozen. This ensures the projected audio tokens carry sufficient discriminative information for the language model to distinguish between acoustically different groups, providing the representational foundation for subsequent stages.

Stage~II introduces the episodic format (Section~\ref{sec:episodic_format}) while continuing to train only the encoder and projection. By shifting to few-shot episodes, this stage refines audio representations for comparative reasoning: the encoder must now produce embeddings that are discriminative within the relative context of a support set, not merely separable across a fixed set of clusters.

Stage~III freezes the encoder and projection entirely and applies LoRA adapters~\cite{hu2022lora} to the language model. With the cross-modal interface stabilized, this stage specializes the language model's attention mechanisms for audio-conditioned in-context reasoning while preserving the pretrained medical knowledge encoded in its frozen base weights. The progressive separation of trainable components prevents the language model's large parameter space from dominating early training before audio representations are properly aligned. Table~\ref{tab:ablation} empirically validates the contribution of each stage.
\vspace{-0.15cm}
\subsection{Federated Protocol}
\vspace{-0.15cm}
All three stages execute under federated learning environment implemented using Flower\cite{beutel2020flower}. Clients train on local data and synchronize trainable parameters via FedProx~\cite{li2020federated} at a central server after each round. The communicated parameters are the encoder and projection in Stages~I--II, and LoRA weights in Stage~III; pretrained language model weights are never transmitted, reducing communication cost and preserving medical knowledge. Pseudo-labeling is performed independently within each client using only local embeddings, requiring no shared label taxonomy across institutions. Raw audio never leaves its source institution throughout the entire training process.

\section{Experiments}
\subsection{Datasets}
\label{sec:datasets}
\vspace{-0.1cm}

We evaluate on seven medical audio datasets spanning respiratory and cardiac domains (Table~\ref{tab:datasets}). In the federated setup, each dataset is assigned to a separate client, yielding $M{=}7$ institutions with naturally heterogeneous recording conditions, label spaces, and sample sizes. All audio is resampled to 16\,kHz and segmented into fixed-length clips of 5 seconds with zero-padding for shorter recordings. Train splits are used for federated training with pseudo-labels; held-out test splits are used exclusively for episodic evaluation with real clinical labels.

\begin{table}[t]
  \centering
  \vspace{-0.3cm}

  \caption{Summary of federated client datasets spanning respiratory and cardiac audio domains.}
  \label{tab:datasets}
  \small
  \begin{adjustbox}{max width=\columnwidth}
  \begin{tabular}{llcc}
    \toprule
    Dataset & Domain & Train & Test \\
    \midrule
    ICBHI~\cite{rocha2019open}         & Respiratory sounds        & 539 & 384 \\
    CIRCOR~\cite{PhysioNet-circor-heart-sound-1.0.3}       & Heart sounds              & 1906 & 499 \\
    CoughVID~\cite{orlandic2021coughvid}   & Cough recordings          & 3527 & 64  \\
    HFLUNG~\cite{hsu2021benchmarking}       & Lung sounds               & 5663 & 465 \\
    SPRSound~\cite{zhang2022sprsound}   & Respiratory sounds        & 1256  & 154 \\
    COVID-19 Sounds~\cite{xia2021covid} & COVID-19 cough/breath     & 4000 & 1243 \\
    ZCHSound~\cite{jia2024zchsound}   & Heart sounds              & 588  & 174 \\
    \midrule
    \textbf{Total} & Cardiopulmonary & \textbf{17479} & \textbf{2983} \\
    \bottomrule
  \end{tabular}
  \end{adjustbox}
  \vspace{-0.5cm}
\end{table}

\begin{table*}[htbp]
\centering
\vspace{-0.2cm}
\caption{Comparison with baseline methods on episodic evaluation (1,000 episodes, 5 seeds). Results are averaged over $14$ categories. Accuracy (\%), ROUGE-L, and BERTScore F1 (\%) are reported as mean ± std.}
\label{tab:baseline_comparison}
\vspace{-0.1cm}
\small
\resizebox{\textwidth}{!}{
\begin{tabular}{lcccccccccccc}
\toprule
\multirow{2}{*}{Method} & \multicolumn{3}{c}{2-way-2-shot} & \multicolumn{3}{c}{2-way-5-shot} & \multicolumn{3}{c}{3-way-2-shot} & \multicolumn{3}{c}{3-way-5-shot} \\
\cmidrule(lr){2-4} \cmidrule(lr){5-7} \cmidrule(lr){8-10} \cmidrule(lr){11-13}
& Acc. & ROUGE-L & BERT & Acc & ROUGE-L & BERT & Acc. & ROUGE-L & BERT & Acc. & ROUGE-L & BERT \\
\midrule
Pengi \cite{deshmukh2023pengi} & 50.91±2.13 & 52.34±1.37 & 54.18±1.56 & 51.50±1.23 & 53.12±0.76 & 55.01±0.84 & 32.68±1.55 & 33.45±1.81 & 33.32±3.01 &  35.43±0.45 & 36.39±1.29 & 36.85±1.35 \\
Gama \cite{ghosh2024gama} & 51.33±1.56 & 52.45±2.13 & 53.26±1.87 & 50.40±0.87 & 51.34±1.82 & 52.76±2.20 & 34.00±2.13 & 35.40±1.24 & 38.51±1.62 & 36.15±1.51 & 37.23±0.97 & 39.12±1.26 \\
Gemma3N \cite{team2025gemma} & 44.03±1.37 & 47.96±1.42 & 50.66±1.28 & 26.70±1.08 & 30.27±1.23 & 30.35±0.94 & 32.30±1.17 & 36.68±1.31 & 40.56±0.89 & 23.60±1.05 & 28.02±1.54 & 30.39±1.46 \\
Qwen2.5-Omni-7B \cite{Qwen2.5-Omni} & 62.07±0.52 & 65.35±0.47 & 67.03±0.35 & 63.00±0.49 & 66.45±0.48 & 67.96±0.17 & 48.20±0.27 & 51.70±0.57 & 55.41±1.30 & 44.90±1.54 & 49.76±1.63 & 52.94±1.76 \\
Audio Flamingo 3 \cite{goel2025audio} & 29.40±2.52 & 30.40±2.41 & 25.74±2.86 & 41.60±1.970 & 30.04±2.32 & 24.02±1.83 & 44.50±1.30 & 29.01±1.56 & 25.42±1.74 & 30.10±3.21 & 18.32±2.50 & 11.42±2.24 \\

\midrule
\textbf{FSC (Ours)} & \textbf{71.61±1.51} & \textbf{73.72±1.41} & \textbf{75.59±0.88} & \textbf{68.34±1.24} & \textbf{70.76±0.81} & \textbf{73.14±0.85} & \textbf{54.29±3.29} & \textbf{56.18±2.93} & \textbf{59.83±2.77} & \textbf{51.78±1.24} & \textbf{52.47±1.28} & \textbf{55.68±1.02} \\
\bottomrule
\end{tabular}
}

\end{table*}

\begin{table*}[htbp]
\centering
\vspace{-0.2cm}
\caption{Class-wise and overall few-shot diagnostic performance of FSC across episodic configurations, averaged over 1{,}000 episodes and 5 seeds.}
\vspace{-0.2cm}
\small
\label{tab:fake_label_federated_episodic}
\resizebox{\textwidth}{!}{
\begin{tabular}{lcccccccccccc}
\toprule
\multirow{2}{*}{Category} & \multicolumn{3}{c}{2-way-2-shot} & \multicolumn{3}{c}{2-way-5-shot} & \multicolumn{3}{c}{3-way-2-shot} & \multicolumn{3}{c}{3-way-5-shot} \\
\cmidrule(lr){2-4} \cmidrule(lr){5-7} \cmidrule(lr){8-10} \cmidrule(lr){11-13}
& Acc. & ROUGE-L & BERT & Acc. & ROUGE-L & BERT & Acc. & ROUGE-L & BERT & Acc. & ROUGE-L & BERT \\
\midrule
URTI & 89.25±6.95 & 89.92±6.38 & 90.44±6.18 & 90.42±5.15 & 90.89±4.85 & 91.26±4.84 & 74.99±5.87 & 76.75±4.78 & 78.95±4.63 & 60.94±5.64 & 66.21±4.27 & 68.52±3.23 \\
Heart Sound Abnormal & 80.68±5.39 & 82.95±6.08 & 80.17±6.58 & 85.33±4.67 & 86.59±4.18 & 84.08±3.77 & 64.16±1.20 & 63.42±2.43 & 59.45±1.23 & 55.78±3.32 & 59.43±2.54 & 52.30±2.91 \\
Mixed Respiratory Abnormal & 75.42±7.26 & 77.37±6.64 & 80.64±6.12 & 76.26±3.27 & 77.81±3.27 & 81.25±2.77 & 38.89±2.24 & 40.89±1.60 & 47.9±4.74 & 27.19±9.31 & 31.20±9.24 & 38.51±7.38 \\
Atrial Septal Defect & 74.92±2.97 & 76.88±2.91 & 78.22±2.64 & 72.78±2.91 & 75.11±2.72 & 76.33±2.76 & 73.42±2.92 & 74.84±3.19 & 76.33±2.52 & 66.20±3.90 & 67.23±4.28 & 67.28±4.09 \\
Bronchiolitis & 74.79±4.09 & 77.81±4.82 & 80.06±4.17 & 72.51±7.67 & 75.63±6.88 & 78.94±6.37 & 60.47±2.56 & 63.69±2.13 & 68.06±1.31 & 57.69±8.43 & 60.16±7.98 & 63.94±5.60 \\
Wheeze & 74.32±3.03 & 77.19±2.61 & 80.63±2.27 & 70.25±3.30 & 73.03±2.55 & 76.76±2.61 & 57.71±3.13 & 58.17±2.35 & 64.29±1.87 & 57.84±3.50 & 58.13±2.07 & 64.13±1.97 \\
Crackle & 70.64±8.16 & 72.98±7.42 & 78.05±6.38 & 71.72±2.12 & 74.11±1.03 & 79.18±0.50 & 43.14±6.15 & 49.70±5.48 & 55.30±5.71 & 56.73±9.49 & 63.67±9.51 & 67.96±8.45 \\
Ventricular Septal Defect & 69.95±2.25 & 73.16±1.53 & 72.68±0.99 & 56.98±3.65 & 58.91±3.54 & 59.02±3.52 & 60.23±2.13 & 62.15±1.55 & 60.55±3.67 & 62.67±6.22 & 62.67±6.22 & 60.29±5.34 \\
Respiratory Healthy & 69.35±4.98 & 71.12±4.50 & 75.10±4.06 & 59.57±9.64 & 62.26±8.98 & 68.77±8.29 & 51.27±8.67 & 56.86±6.94 & 61.28±7.35 & 45.38±1.52 & 51.73±1.99 & 57.68±1.59 \\
Heart Healthy & 67.04±4.45 & 69.83±4.14 & 71.05±2.75 & 58.54±3.25 & 61.27±2.92 & 63.42±2.35 & 55.00±5.24 & 57.91±5.02 & 61.43±4.74 & 46.21±5.71 & 48.95±5.29 & 53.48±4.74 \\
Stridor & 65.53±4.31 & 66.36±3.77 & 68.09±3.12 & 69.45±5.90 & 71.22±5.14 & 73.81±4.44 & 58.68±8.52 & 54.39±5.23 & 61.27±3.80 & 44.51±5.16 & 44.12±6.62 & 53.13±6.50 \\
COVID & 65.49±2.07 & 67.96±2.40 & 71.60±1.95 & 62.50±1.38 & 67.48±1.59 & 70.44±1.59 & 42.52±7.34 & 44.06±7.18 & 49.39±7.44 & 39.39±7.99 & 40.67±5.39 & 45.25±7.97 \\
Rhonchi & 63.25±2.95 & 66.12±3.27 & 69.58±3.46 & 60.41±2.96 & 63.98±3.02 & 67.57±2.52 & 45.98±2.22 & 46.88±1.26 & 42.67±1.48 & 41.06±5.12 & 43.76±2.30 & 42.06± 1.20 \\
COPD & 61.25±3.38 & 62.58±3.48 & 63.38±2.74 & 54.38±4.78 & 55.35±4.71 & 56.16±5.05 & 37.35±5.42 & 44.79±4.42 & 48.12±3.76 & 33.96±7.73 & 37.03±5.13 & 40.89±6.64 \\
\midrule
\textbf{FSC (Ours)} & \textbf{71.61±1.51} & \textbf{73.72±1.41} & \textbf{75.59±0.88} & \textbf{68.34±1.24} & \textbf{70.76±0.81} & \textbf{73.14±0.85} & \textbf{54.29±3.29} & \textbf{56.18±2.93} & \textbf{59.83±2.77} & \textbf{51.78±1.24} & \textbf{52.47±1.28} & \textbf{55.68±1.02} \\
\bottomrule
\end{tabular}
}

\end{table*}
\vspace{-0.1cm}

\vspace{-0.1cm}
\subsection{Experimental Setup}
\label{sec:setup}
\vspace{-0.1cm}
\noindent\textbf{Evaluation protocol.}\quad
We adopt an $N$-way $K$-shot episodic protocol with $N \in \{2,3\}$ and $K \in \{2,5\}$. For each episode, one query sample is drawn first; its support set is then constructed by sampling $K$ examples from each of $N$ randomly selected clinical categories. To evaluate cross-domain generalization, episodes mix samples across datasets so that support and query may originate from different institutions. All samples are drawn from held-out test splits. Results are averaged over 1{,}000 episodes across five random seeds.

\noindent\textbf{Metrics.}\quad
We report Accuracy for exact label match, ROUGE-L F1~\cite{lin2004rouge} for lexical overlap, and BERTScore F1~\cite{zhang2019bertscore} for semantic alignment between generated and ground-truth labels.

\noindent\textbf{Training details.}\quad
We train with the AdamW optimizer~\cite{loshchilov2017decoupled} using a learning rate of $5{\times}10^{-4}$ for Stages~I--II and $2{\times}10^{-4}$ for Stage~III. Stage~I runs for 10 federated rounds, Stage~II for 15 rounds, and Stage~III for 10 rounds. LoRA adapters in Stage~III use rank $r{=}8$ and $\alpha{=}32$. All stages use FedProx with proximal coefficient $\mu{=}0.01$.

\noindent\textbf{Baselines.}\quad
We compare against recent audio-language models: Pengi~\cite{deshmukh2023pengi}, GAMA~\cite{ghosh2024gama}, Gemma3n~\cite{team2025gemma}, Qwen2.5-Omni-7B~\cite{Qwen2.5-Omni}, and Audio Flamingo~\cite{goel2025audio}. All baselines are evaluated under the same episodic protocol by formatting episodes as in-context prompts for each model. None of these baselines are trained in a federated setting, providing an upper-bound comparison against centralized models with full data access.

\section{Results and Analysis}
\label{sec:results}

We evaluate FSC under the label-unseen episodic protocol described in Section~\ref{sec:setup}, where no diagnostic labels are seen during training and classification relies entirely on in-context conditioning at inference. To mitigate linguistic bias from disease phrasing, each condition is represented by three physician-validated textual descriptions and performance is averaged to obtain condition-level results. Tables~\ref{tab:baseline_comparison} and~\ref{tab:fake_label_federated_episodic} report overall and class-wise results.

\begin{table}[t]
\centering
\vspace{-0.3cm}
\caption{Ablation study of FSC components under the 2-way 2-shot setting (averaged over 1{,}000 episodes).}
\label{tab:ablation}
\vspace{-0.2cm}
\small
\setlength{\tabcolsep}{5pt}
\renewcommand{\arraystretch}{1.15}
\begin{adjustbox}{max width=\linewidth}
\begin{tabular}{@{}lccc@{}}
\toprule
\textbf{Configuration} & \textbf{Acc.} & \textbf{ROUGE-L} & \textbf{BERTScore} \\
\midrule
Full FSC (Ours)     & \textbf{71.61} & \textbf{73.72} & \textbf{75.59} \\
\midrule
\multicolumn{4}{@{}l}{\textit{Training strategy}} \\
\quad Joint training (Proj.\ + LoRA)        & 65.87 & 66.84 & 68.08 \\
\quad w/o Stage~I alignment            & 67.31 & 69.18 & 71.89 \\
\midrule
\multicolumn{4}{@{}l}{\textit{Input modality}} \\
\quad w/o audio embeddings            & 49.68 & 53.65 & 58.38 \\
\midrule
\multicolumn{4}{@{}l}{\textit{Aggregation and data setting}} \\
\quad FedAvg~\cite{mcmahan2017communication}                    & 69.63 & 71.58 & 73.53 \\
\quad Centralized + pseudo-labels       & 65.44 & 66.95 & 70.12 \\
\midrule
\multicolumn{4}{@{}l}{\textit{Language model backbone}} \\
\quad LLaMA3.2-1B~\cite{grattafiori2024llama}  & 49.23 &     51.93 &    52.89 \\
\quad Qwen2.5-1.5B~\cite{qwen2.5}      & 50.32& 53.30 &57.72 \\
\midrule
\multicolumn{4}{@{}l}{\textit{Number of pseudo-label clusters $C$}} \\
\quad $C{=}2 / 4 / 16$   & 57.46 / 57.72 / 65.36 & 61.92 / 60.96 / 68.31 & 63.64 / 64.33 / 69.51 \\
\quad $C{=}24 / 32$       & 60.36 / 56.18          & 63.31 / 59.69          & 66.51 / 63.08          \\
\bottomrule
\end{tabular}
\end{adjustbox}
\vspace{-0.7cm}
\end{table}

\noindent\textbf{Comparison with baselines.}\quad
FSC consistently outperforms all audio-language baselines across every episodic configuration (Table~\ref{tab:baseline_comparison}). In the 2-way 2-shot setting, FSC achieves 71.6\% accuracy, exceeding the strongest baseline by over 9 percentage points (71.6\% vs.\ 62.1\%). This margin holds across all three metrics, with ROUGE-L and BERTScore improvements of comparable magnitude, confirming that the gains reflect genuine diagnostic capability rather than surface-level lexical matching. The gap is especially notable given that all baselines are centralized models with unrestricted data access, whereas FSC trains under federated constraints with no real labels.

\noindent\textbf{Generalization across clinical domains.}\quad
Class-wise results (Table~\ref{tab:fake_label_federated_episodic}) reveal strong performance on conditions with distinctive acoustic signatures across both respiratory and cardiac domains. For respiratory diagnosis, FSC achieves 89.3\% on URTI, 74.8\% on bronchiolitis, 74.3\% on wheeze, and 70.6\% on crackle in the 2-way 2-shot setting, indicating effective capture of characteristic patterns associated with airway obstruction and abnormal breath sounds. Cardiac conditions show comparable accuracy: 80.7\% for abnormal heart sounds, 74.9\% for atrial septal defect, and 70.0\% for ventricular septal defect, demonstrating sensitivity to pathological murmurs and irregular rhythms. This cross-domain consistency confirms that FSC acquires generalizable audio-language representations rather than domain-specific shortcuts, and that the pseudo-label training strategy transfers effectively to real clinical terminology in both organ systems.

\noindent\textbf{Effect of episode configuration.}\quad
FSC maintains strong 2-way performance (71.6\% for 2-shot, 68.3\% for 5-shot), while 3-way settings show expected degradation (54.3\% and 51.8\%) as discriminating among more categories within limited context becomes harder. Within the same $N$-way setting, increasing from 2-shot to 5-shot yields a slight accuracy decrease rather than the improvement typically seen in metric-learning few-shot methods. We attribute this to the autoregressive nature of the model: longer multimodal input sequences extend the context window and can dilute attention over critical acoustic features, an effect not present in embedding-distance classifiers. Peak performance occurs when the evaluation format matches the training episode configuration, underscoring the importance of episodic consistency between training and inference.

\noindent\textbf{Ablation analysis.}\quad
Table~\ref{tab:ablation} isolates each design choice under the 2-way 2-shot setting. Removing Stage~I alignment drops accuracy by 4.3 points, and jointly training the projection and LoRA from scratch degrades it by 5.7 points, validating the progressive pipeline: the language model benefits from a stabilized audio interface before its own parameters are adapted. Zeroing audio embeddings at inference causes accuracy to fall near chance (49.7\%), confirming that FSC genuinely relies on acoustic content rather than exploiting language-only priors from MedGemma. On the federated side, FedProx outperforms FedAvg by 2 points (71.6\% vs.\ 69.6\%), reflecting its effectiveness at mitigating client drift under the heterogeneous data distributions present across the seven institutional clients. Most notably, federated training surpasses centralized training under the same pseudo-label strategy by over 6 points (71.6\% vs.\ 65.4\%). We attribute this to the natural regularization effect of local optimization across diverse institutional distributions: each client develops representations shaped by its own recording conditions and pathology mix, and aggregation encourages features that generalize across these variations rather than overfitting to a single mnolithic clustering structure. Replacing MedGemma with general-purpose LLMs (LLaMA3.2-1B: 49.2\%, Qwen2.5-1.5B: 50.3\%) leads to substantial drops, confirming the advantage of medically pretrained language priors.

\section{Conclusions}
\label{sec:conclusions}
\vspace{-0.1cm}
This paper introduced FSC, a federated framework for few-shot clinical audio diagnosis that operates without centralized data sharing or real diagnostic labels. FSC pairs pseudo-label episode construction with a three-stage training pipeline to equip a multimodal language model with in-context diagnostic reasoning from limited support examples. Experiments on held-out respiratory and cardiac conditions show that FSC surpasses all audio-language baselines by a substantial margin. Notably, decentralized training proves more effective than its centralized counterpart under identical labeling conditions, suggesting that cross-institutional diversity acts as a natural regularizer. These results establish that privacy-preserving few-shot diagnosis of clinical audio is practically viable. Future directions include reducing sensitivity to episode format at inference time and extending the framework to other speech-based clinical assessments such as neurological screening.

\clearpage
\section{Generative AI Use Disclosure}
\small{Generative AI tools were used solely for language editing and polishing to improve clarity and readability of the manuscript. All technical content, experimental design, analysis, and conclusions were created by the authors. The authors take full responsibility for the content of this paper.}

\section{Acknowledgments}
\small{This work was supported by the NWO AiNed Fellowship Grant of A.S., and in part by Google.org and the Google Cloud Research Credits program through the Gemini Academic Program. We also acknowledge the use of the Dutch National Supercomputer Snellius for essential computational tasks.}

\bibliographystyle{IEEEtran}
\bibliography{mybib}

\end{document}